\documentclass[11pt]{article}

\usepackage[preprint]{acl}

\usepackage{times}
\usepackage{latexsym}
\usepackage{hyperref}  %
\usepackage[T1]{fontenc}
\usepackage[utf8]{inputenc}
\usepackage{microtype}

\usepackage{inconsolata}
\usepackage{amsmath, amssymb, amsfonts}
\usepackage{graphicx}
\usepackage{tcolorbox}
\usepackage{graphicx}
\usepackage{algorithm}
\usepackage{algorithmic} 
\usepackage{booktabs}
\usepackage{multirow}
\usepackage{makecell}
\usepackage{tcolorbox}
\usepackage{xcolor}
\usepackage{lipsum} % 仅用于示例文本，实际使用时可以删除
\definecolor{bluebar}{RGB}{0, 120, 215}
\definecolor{lightblue}{RGB}{240, 248, 255}
\title{Omni-AutoThink: Adaptive Multimodal Reasoning via Reinforcement Learning}

\author{
 \textbf{Dongchao Yang\textsuperscript{1}},
 \textbf{Songxiang Liu\textsuperscript{2,\textdagger}},
 \textbf{Disong Wang\textsuperscript{2}},
 \textbf{Yuanyuan Wang\textsuperscript{1}},
\\
 \textbf{Guanglu Wan\textsuperscript{2}},
 \textbf{Helen Meng\textsuperscript{1}}
\\
\\
 \textsuperscript{1} The Chinese University of Hong Kong,
 \textsuperscript{2} Meituan
\\
 \small{
   \textbf{\textsuperscript{\textdagger} Correspondence:} \href{mailto:email@domain}{songxiangliu.cuhk@gmail.com}
 }
}

\begin{document}
\maketitle
\begin{abstract}
Recent advances in Omni models have enabled unified multimodal perception and generation. However, most existing systems still exhibit rigid reasoning behaviors—either overthinking simple problems or failing to reason when necessary.
To address this limitation, we propose Omni-AutoThink, a novel adaptive reasoning framework that dynamically adjusts the model's reasoning depth according to task difficulty.
Our framework comprises two stages: (1) an Adaptive Supervised Fine-Tuning (Adaptive SFT) stage, which endows the Omni model with fundamental reasoning capability using large-scale reasoning-augmented data, and (2) an Adaptive Reinforcement Learning (Adaptive GRPO) stage, which optimizes reasoning behaviors based on task complexity and reward feedback.
We further construct a comprehensive adaptive reasoning benchmark that spans text-only, text–audio, text–visual, and text–audio–visual modalities, providing both training and evaluation splits for multimodal reasoning assessment.
Experimental results demonstrate that our proposed framework significantly improves adaptive reasoning performance compared to previous baselines.
All benchmark data and code will be publicly released at: \url{ https://github.com/yangdongchao/Omni-AutoThink}.

\end{abstract}

\begin{figure*}[t]
    \centering
    \includegraphics[width=0.98\textwidth]{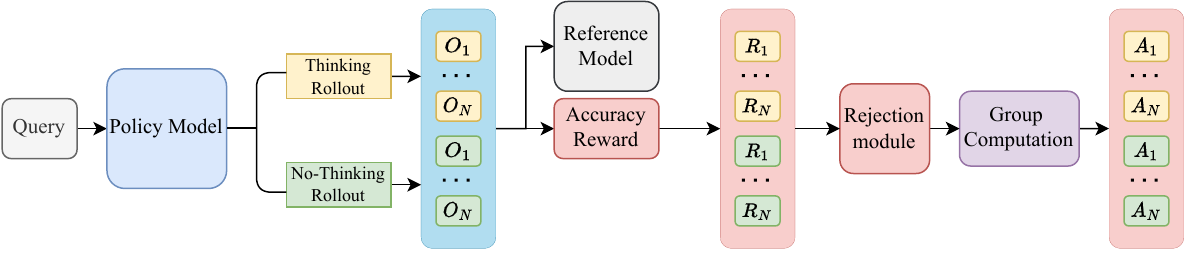}
    \caption{The overview of proposed adaptive GRPO. The proposed method follows the setting of GRPO. We introduce the adaptive sampling strategy during the training stage, refer to Section \ref{sub_sec:adaptive grpo} find the details.}
    \label{fig:overview}
\end{figure*}

\section{Introduction}
The success of Large Language Models (LLMs), such as GPT-4 \citep{gpt4} and LLaMA \citep{llama}, has inspired the rapid development of Multimodal Large Language Models (MLLMs), include audio-language models \citep{qwen-audio,tang2023salmonn}, vision-language models \citep{minigpt5,liu2024language,yu2024spae,zhu2024beyond,li2023blip,tsimpoukelli2021multimodal}, and Omni models \citep{qwen3-omni,gemini,LongCat-Flash-Omni}. Recently, Large Reasoning Models (LRMs), such as GPT-O1 and DeepSeek-R1 \citep{deepseek-r1}, have further advanced the reasoning capability of LLMs in solving complex problems.
% The core idea of these reasoning models is to initiate a deliberate thinking process before predicting the final answer.
Similarly, step-by-step reasoning has shown substantial benefits in MLLMs.
For example, Qwen3-Omni-think improves the performance on complex audio understanding tasks by reasoning, while Vision-R1 \cite{vision-r1} demonstrates that enhancing reasoning ability in vision-language models significantly boosts visual comprehension.

However, existing Omni models typically operate in one of two rigid modes:
(1) a thinking mode, where the model produces a reasoning trace for every query—leading to unnecessary computational overhead on simple problems; or
(2) a no-thinking mode, where the model directly outputs answers without reasoning—resulting in poor performance on complex problems.
This phenomenon motivates an auto-thinking paradigm, where the model automatically decides whether to engage in reasoning based on the complexity of the input.

Although several pioneering works have explored automatic reasoning control \citep{adaptthink,r-4b,audio-thinker}, limitations in adaptive Omni reasoning still persist:
(1) some approaches depend on manual configurations (e.g., Qwen3 \citep{qwen3});
(2) others \citep{kat-v1,adaptthink} rely on handcrafted data or intricate reward functions during reinforcement learning; and
(3) prior studies \citep{adaptthink,kat-v1,audio-thinker,r-4b} are restricted to limited modalities(e.g. text-only, text-audio, or text-vision) modalities.

In this study, we introduce Omni-AutoThink, an Omni model designed for automatic reasoning across any modality, including text-only, text-vision, text-audio, and text-audio-vision scenarios.
Omni-AutoThink can dynamically choose between reasoning and non-reasoning modes depending on the difficulty of the user’s query.
To achieve this goal, we propose a two-stage training framework consisting of:
(1) an Adaptive Supervised Fine-Tuning (SFT) warm-up stage, and
(2) an Adaptive Reinforcement Learning stage.
To support SFT, we design a data construction strategy to build a large-scale adaptive reasoning dataset across multiple modalities.
% The key idea is to use a teacher model to distinguish between easy and difficult problems, thereby generating reasoning traces selectively.
However, we observe that the model tends to collapse into either always reasoning or never reasoning—failing to achieve adaptive control after SFT training. 
To address this issue, we introduce Adaptive GRPO, a reinforcement learning algorithm tailored for auto-thinking based on GRPO \citep{deepseek-r1}.
Unlike previous complex RL methods \citep{adacot,adaptthink,audio-thinker}, our Adaptive GRPO relies solely on accuracy-based rewards.
During training, we sample both reasoning and non-reasoning trajectories for each problem: the model is encouraged to answer directly for easy problems, while reasoning is incentivized for difficult ones.
Through this process, the model gradually learns when to think and when not to think.

Furthermore, the research community currently lacks a unified benchmark to evaluate Omni auto-thinking performance.
To fill this gap, we propose AdaptiveBenchmark, which covers multiple modalities and provides detailed difficulty annotations.
Extensive experiments demonstrate that our proposed model achieves superior adaptive reasoning performance compared to previous methods.

\section{Related Works}
\subsection{Multimodal Large Language Models}
As Large Language Models (LLMs) continue to advance rapidly, they have paved the way for the evolution of Multimodal Large Language Models (MLLMs) \citep{audio-flamingo,qwen3-omni,LongCat-Flash-Omni,uniaudio1.5}.
The development of MLLMs has undergone two major stages, progressing from unimodal approaches—such as audio-language models \citep{audio-flamingo,tang2023salmonn,uniaudio1.5} and vision-language models \citep{alayrac2022flamingo,minigpt5,li2023blip}—to omni-modal approaches represented by Gemini \citep{gemini}, Qwen2.5-Omni \citep{qwen3-omni}, and LongCat-Omni \citep{LongCat-Flash-Omni}.
We observe that the evolution of MLLMs is characterized by three major trends:
(1) the pursuit of unified architectures capable of handling diverse modalities within a single framework;
(2) the transition from pure perception capabilities toward cognitive reasoning; and
(3) an increasing focus on engineering optimization and deployment efficiency for real-world applications.
These developments collectively push the boundaries of what multimodal systems can achieve, moving beyond simple cross-modal understanding toward comprehensive world modeling and interactive intelligence.
In this work, we focus on improving the reasoning ability of Omni models.

\subsection{Adaptive Multimodal Reasoning}
The pursuit of efficient reasoning in large language models has evolved through several paradigms.
Early approaches focused on fixed reasoning patterns, where models followed predetermined reasoning templates regardless of problem complexity \citep{qwen3}.
More recent works have explored adaptive reasoning strategies, allowing models to decide whether to engage in a reasoning process before answering a user’s query \citep{adacot,adaptthink}.
In multimodal contexts, researchers have proposed several difficulty-aware reasoning methods that adjust computational budgets according to the perceived complexity of a problem \citep{audio-thinker,r-4b}.

Inspired by these approaches, our method introduces a unified framework for adaptive reasoning in Omni models, which can automatically adjust its reasoning behavior across any modality and difficulty level—without requiring explicit triggers or manually defined reasoning modes.

\section{Methods}
% 写作逻辑
% problem formulation
% discussion the obervation
% our solution
\subsection{Problem Formulation}
We define \textbf{adaptive reasoning} as the ability of an omni-modal model to flexibly determine \textit{when} and \textit{how deeply} to reason in response to varying task difficulty.
Let $\pi_\theta$ denote a multimodal reasoning policy parameterized by $\theta$, mapping an input query $x$ to a generated response $y$.
Unlike conventional reasoning models that perform reasoning with a fixed depth, an adaptive reasoning model aims to \textit{self-regulate} the extent of reasoning under different input conditions.
Formally, the response generation follows an autoregressive process:
\begin{equation}
\pi_\theta(y|x) = \prod_{t=1}^{m} \pi_\theta(y_t \mid x, y_{<t}),
\label{eq:autoregressive}
\end{equation}
where $y_t$ is the token emitted at step $t$. 
We divide the response $y$ into two conceptual components:
\begin{equation}
y = (y^{\text{reason}}, y^{\text{answer}}),
\end{equation}
where $y^{\text{reason}}$ denotes the internal reasoning trace (within \texttt{<think>} ... \texttt{</think>}) and $y^{\text{answer}}$ is the final output that solves the task. 
The model’s challenge lies in learning when to activate $y^{\text{reason}}$.

Let $\mathcal{D} = \{(x_i, y_i)\}_{i=1}^N$ denote a multimodal training corpus containing text, audio, and visual modalities. 
Each example has an associated \textit{difficulty indicator} $d_i \in [0,1]$, which reflects the complexity of the task. 
The adaptive reasoning objective seeks to optimize both the \textit{task performance} and the \textit{reasoning efficiency}, defined as:
\begin{equation}
\begin{aligned}
\max_\theta \ 
\mathbb{E}_{(x_i, y_i) \sim \mathcal{D}} 
\Big[ R_\text{t}(x_i, y_i; \pi_\theta)
- \lambda(d_i)C_\text{r}(x_i; \pi_\theta) \Big].
\end{aligned}
\label{eq:adaptive_obj}
\end{equation}
where $R_\text{t}$ measures task success (e.g., accuracy), 
$C_\text{r}$ quantifies the cost of reasoning (e.g., CoT length), 
and $\lambda(d_i)$ is a dynamic weighting coefficient that adjusts the trade-off between accuracy and reasoning effort according to task difficulty $d_i$.

Intuitively, the model learns to reason \textit{more} for challenging tasks and \textit{less} for trivial ones, thus achieving adaptive control of reasoning depth. 
This formulation provides the foundation for our proposed \textbf{Omni-AutoThink} framework, which unifies adaptive reasoning learning under both supervised and reinforcement objectives.

\begin{table}[ht]
\small
\centering
\caption{Preliminary studies on adaptive reasoning strategies. We conduct experiments on text-audio (audio question answer) tasks. 
Omni+RL denotes applying GRPO to encourage the model to adaptively reason using both format and accuracy rewards.
Omni+RL (No) denotes training the model without requiring it to output the reasoning process, where optimization is guided solely by the accuracy reward.}
\label{tab:preliminary_study}
\begin{tabular}{l*{4}{c}}
\toprule
\multirow{2}{*}{Model} & \multicolumn{2}{c}{Easy} & \multicolumn{2}{c}{Hard} \\
\cmidrule(lr){2-3} \cmidrule(lr){4-5}
 & Accuracy & Rate & Accuracy & Rate \\
\midrule
Base prompt & 0.98 & 0.00 & 0.29 & 0.00 \\
Adaptive prompt & 0.94 & 0.01 & 0.34 & 0.01 \\
\midrule
Omni + SFT & 0.95 & 0.00 & 0.42 & 0.00 \\
\midrule
Omni + RL & 0.98 & 0.00 & 0.45 & 0.00 \\
Omni + RL (No) & 0.99 & 0.00 & 0.48 & 0.00 \\
\bottomrule
\end{tabular}
\end{table}

\subsection{Discussion the observations}
To explore the feasibility of achieving adaptive reasoning, we conduct a series of preliminary studies based on three intuitive assumptions. Each assumption reflects a potential training or prompting strategy, followed by our empirical observations and analysis.

\noindent \textbf{Assumption 1: Prompt-based adaptive reasoning.}
We hypothesize that an omni-modal model can self-assess problem difficulty through appropriate prompting:
if the problem is easy, it should output the answer directly; otherwise, it should first generate a reasoning process before producing the final answer.

\textbf{Observation.}
Using Qwen2.5-Omni-7B as the base model and the prompt design shown in Appendix~\ref{appendix:prompt_design}, we evaluate its performance on the audio question answer test set.
As shown in Table \ref{tab:preliminary_study}, the model fails to exhibit adaptive reasoning behavior: it almost never reason for both easy and hard problem: indicating that prompting alone is insufficient.
This suggests that adaptive reasoning is a non-trivial ability that cannot simply be induced by external instructions.

\noindent \textbf{Assumption 2: Supervised fine-tuning (SFT) on difficulty-annotated data.}
We next consider whether SFT on a dataset containing both easy and hard samples can lead to adaptive reasoning.
In this dataset, easy problems include only direct answers, while hard problems contain both reasoning traces and final answers.

\textbf{Observation.}
We fine-tune Qwen2.5-Omni-7B using the difficulty-annotated dataset.
As shown in Table \ref{tab:preliminary_study}, the fine-tuned model fails to develop adaptive reasoning and instead collapses into extreme behavior: the \textit{no-thinking mode} (never reasoning). 
This finding implies that simple supervised signals are inadequate to teach the model when to think.

\noindent \textbf{Assumption 3: Reinforcement learning with format and accuracy rewards.} 
We further test whether a reinforcement learning algorithm, such as GRPO~\citep{deepseek-r1}, can induce adaptive reasoning.
The same dataset from the SFT experiment is used. We follow the setting of AdaptCoT \cite{adacot} and Audio-Thinker \cite{audio-thinker}, both the format and accuracy rewards are set to 1. 

\textbf{Observation.}
As shown in Table \ref{tab:preliminary_study}, directly applying GRPO causes the model to collapse into the no-thinking mode, producing no reasoning traces for any problem.
We identify two potential causes:
(1) the pre-defined easy/hard labels become invalid as the model's capability improves during training, making the fixed format reward ineffective;
and (2) we observe, consistent with prior studies~\citep{r1-vqa,audio-thinker,omni-r1}, that the model achieves higher accuracy when reasoning is omitted during RL training (as Table \ref{tab:preliminary_study} row 4-5 shows).
This leads to a biased optimization toward non-reasoning behavior.
We attribute this to \textbf{the base model’s lack of inherent reasoning ability, as Qwen2.5-Omni was not exposed to sufficient reasoning traces during pre-training.}

\noindent \textbf{Summary.}
Across all three assumptions, our results consistently reveal that achieving adaptive reasoning is far from trivial.
Prompt-based control lacks effectiveness, SFT collapses into fixed reasoning modes, and simple RL-based methods easily degenerate toward non-reasoning behavior.
Although several prior works~\citep{adacot,audio-thinker} have achieved partial success using complex reinforcement learning frameworks, these models are closed-source and confined to text-only or audio-only domains.

\noindent In the next section, we present the details of the proposed \textbf{Omni-AutoThink}, including the adaptive SFT and Adaptive GRPO. 

\subsection{Adaptive SFT} \label{sub_sec:adaptive sft}
The Supervised Fine-Tuning (SFT) stage primarily focuses on two objectives:
(1) enhancing the base model’s reasoning capability by introducing large-scale reasoning trajectory data; and
(2) enabling the model to learn both the \textit{thinking} and \textit{no-thinking} modes.

In our study, we define the two modes as follows:
(1) Thinking mode: \texttt{<think>\textbackslash n reasoning step \textbackslash n</think> <answer> answer </answer>} \\
(2) No-thinking mode: \texttt{<think>\textbackslash n\textbackslash n</think> <answer> answer </answer>} \\
As described in Section~\ref{sec:calibration}, manually labeling large-scale datasets as “easy” or “hard” would require significant human effort and computational resources.
Therefore, we construct our SFT dataset in two complementary parts: \\
\textbf{Coarse-level adaptive SFT data}
This component contains large-scale samples, particularly those with reasoning trajectories. As shown in Table~\ref{tab:data_composition}, we do not enforce strict labeling rules to separate easy and hard problems.
Instead, we treat all samples containing reasoning trajectories as \textit{thinking-mode} data, and those without reasoning trajectories as \textit{no-thinking} data.

\noindent \textbf{Precise-level adaptive SFT data}
In this component, we follow the data construction strategy introduced in Section~\ref{sub_sec:benchmark}.
Multiple teacher models are employed to build a smaller, high-quality SFT dataset with explicit difficulty annotations, providing more accurate supervision for adaptive reasoning.

\noindent \textbf{Training Strategy.}
We initialize the model with Qwen2.5-Omni-7B as the base architecture.
Training is conducted in two sequential phases:
(1) we first train the base model for one epoch on the large-scale coarse-level dataset to acquire general reasoning ability; and
(2) we then continue fine-tuning it for one epoch on the precise-level adaptive SFT dataset to refine its discrimination between thinking and no-thinking modes.

\subsection{Adaptive GRPO} \label{sub_sec:adaptive grpo}
The proposed \textbf{Adaptive GRPO} methodology aims to enhance the model’s decision-making capability through direct optimization, enabling it to dynamically select the optimal reasoning strategy for each query.
As illustrated in Figure~\ref{fig:overview}, this approach extends the foundation of \textbf{Group Relative Policy Optimization (GRPO)} by introducing a novel \textit{adaptive rejection sampling} mechanism.
This mechanism ensures a balanced training process by encouraging the model to sample both \textit{thinking} and \textit{no-thinking} trajectories in equal proportion, while strategically masking trivial samples to focus learning on critical decision boundaries. \\
Specifically, Adaptive GRPO introduces two major improvements over vanilla GRPO:
(1) during training, we explicitly force the model to sample both thinking and no-thinking trajectories by manually adjusting the input prompts \citep{qwen3,r-4b}; and
(2) we incorporate a rejection strategy that masks trivial trajectories, thereby emphasizing informative reasoning patterns during optimization. \\
In the following subsections, we first review the definition of GRPO, then present the details of our adaptive sampling strategy and the complete Adaptive GRPO algorithm.

\noindent \textbf{Group Relative Policy Optimization.}
GRPO~\citep{deepseek-r1} is a reinforcement learning algorithm designed to improve policy optimization stability and sample efficiency in large language models.
Unlike traditional approaches such as PPO or DPO that rely on pairwise preference comparisons, GRPO normalizes rewards within a group of sampled responses and computes \textit{relative advantages} across the group.
Given an input query $q$, the algorithm first samples $G$ distinct outputs ${o_1, o_2, \dots, o_G}$.
Each output is evaluated by a reward model $R$, which assigns a scalar reward $r_i$ to each sample.
Based on these rewards, the relative advantage for each output can be computed as shown in Equation~(4).
The full GRPO objective is then summarized in Equation~(5).

\begin{flalign}
\small
&\hat{A}_{i,t} =
\frac{r(q, o_i) - \mathrm{mean}\{r(q,o_1), \ldots, r(q,o_G)\}}
     {\mathrm{std}\{r(q,o_1), \ldots, r(q,o_G)\}} && \tag{4}
\end{flalign}

\begin{flalign}
\small
&\mathcal{L}_{\text{GRPO}i,t} =
\min\Bigg[
\frac{\pi_{\theta}(o_{i,t} \mid q, o_{i,<t})}
     {\pi_{\theta_{\text{old}}}(o_{i,t} \mid q, o_{i,<t})}
     \, \hat{A}_{i,t}, && \nonumber\\[-2pt]
&\qquad
\mathrm{clip}\!\left(
\frac{\pi_{\theta}(o_{i,t} \mid q, o_{i,<t})}
     {\pi_{\theta_{\text{old}}}(o_{i,t} \mid q, o_{i,<t})},
     1-\epsilon,\,
     1+\epsilon
\right)
\hat{A}_{i,t}
\Bigg] && \tag{5}
\end{flalign}
where $\epsilon$ denotes the PPO clipping range.

\noindent \textbf{Adaptive Sampling Strategy.}
From our preliminary experiments, we find that ensuring the sampled trajectories include both \textit{thinking} and \textit{no-thinking} outputs is crucial for stable learning.
Therefore, following the setup of Qwen3~\citep{qwen3}, we design a fixed-prompt mechanism to force the old policy model to sample $G$ thinking outputs and $G$ no-thinking outputs for each query.
Because our old policy model is initialized from the SFT stage, it has already learned both reasoning formats.
To sample a thinking output, we append “\texttt{<think>\textbackslash n}” to the end of the prompt;
to sample a no-thinking output, we append “\texttt{<think>\textbackslash n\textbackslash n</think>}” to the end of the prompt.

\noindent \textbf{Rejection Strategy.}
Under the above adaptive sampling scheme, each query produces $2G$ outputs during training, which inevitably increases computational overhead.
Moreover, we want the model to focus more on difficult problems that are informative for reasoning.
To achieve this, we introduce a \textit{rejection strategy}.
Specifically, for each query, we compute the average reward $r_{\text{avg}}$ across the $2G$ sampled trajectories.
If $r_{\text{avg}}$ exceeds a predefined threshold, we regard the query as an \textit{easy problem} and randomly mask half of its sampled trajectories in both the thinking and no-thinking groups.
This procedure allows the model to allocate greater attention and gradient signal to challenging tasks.

\noindent \textbf{Adaptive GRPO.}
Our Adaptive GRPO builds directly upon vanilla GRPO, differing by the integration of the adaptive sampling and rejection strategies described above.
The core idea is to encourage the model to explore both reasoning modes while relying on the accuracy reward to decide which mode is optimal for each input.
Intuitively, easy problems should be solved without a reasoning process, whereas difficult ones require reasoning to reach correct answers.
The detailed workflow of the adaptive sampling and optimization procedure is summarized in Algorithm~\ref{alg:agrpo_sampling}.

% -------- Adaptive GRPO (ours) --------
\begin{flalign}
\small
& \mathcal{S} = \mathrm{AdaptiveSample}(query),
m_i = \mathbb{I}_{\{\,i \in \mathcal{S}\,\}} && \tag{6}
\end{flalign}
\begin{flalign}
\small
& \hat{A}^{(m)}_{i,t} =
\frac{ r(q,o_i) - \mathrm{mean}\!\{\,r(q,o_{s_1}), \ldots, r(q,o_{s_k})\,\} }
     { \mathrm{std}\!\{\,r(q,o_{s_1}), \ldots, r(q,o_{s_k})\,\} } \, m_i
&& \tag{7}
\end{flalign}
\begin{flalign}
\small
&\mathcal{L}_{\text{AGRPO},i,t} =
\min\Bigg[
\frac{\pi_{\theta}(o_{i,t} \mid q, o_{i,<t})}
     {\pi_{\theta_{\text{old}}}(o_{i,t} \mid q, o_{i,<t})}
     \, \hat{A}^{m}_{i,t}, && \nonumber\\[-2pt]
&\qquad
\mathrm{clip}\!\left(
\frac{\pi_{\theta}(o_{i,t} \mid q, o_{i,<t})}
     {\pi_{\theta_{\text{old}}}(o_{i,t} \mid q, o_{i,<t})},
     1-\epsilon,\,
     1+\epsilon
\right)
\hat{A}^{m}_{i,t}
\Bigg] && \tag{8}
\end{flalign}

\noindent \textbf{Accuracy Reward.}
The reward function is defined in Equation~(9).
Its primary motivation is to encourage the model to solve easy problems without invoking the reasoning process.
By leveraging this accuracy-based reward, the model learns to autonomously decide whether to engage in the \textit{thinking} or \textit{no-thinking} mode according to the complexity of the given problem.

\begin{equation}
R =
\begin{cases}
+2, & \text{if no-think and correct}, \\[3pt]
+1, & \text{if think and correct}, \\[3pt]
0,  & \text{if think and incorrect}, \\[3pt]
-1, & \text{if no-think and incorrect}. \\[3pt]
\end{cases}
\tag{9}
\end{equation}

\begin{algorithm}[t]
%\small
\caption{Adaptive Sampling \& Rejection}
\label{alg:agrpo_sampling}
\begin{algorithmic}[1]
\REQUIRE old policy $\pi_{\text{old}}$, query $q$; group size $G$; threshold $\tau$

\STATE \textbf{(Sampling)} Build prompts:
$\mathcal{P}_{\text{think}}=\{q+ \texttt{<think>\texttt{\textbackslash n}}$,
$\mathcal{P}_{\text{nothink}}=\{q+ \text{<think> \texttt{\textbackslash n} \texttt{\textbackslash n}</think>}$

\STATE Sample $G$ outputs with $\pi_{\text{old}}$ for each type:
$\mathcal{O}_{\text{t}}=\{o_i\}_{i\in I_{\text{t}}}$,
$\mathcal{O}_{\text{n}}=\{o_i\}_{i\in I_{\text{n}}}$, 
where $|I_{\text{t}}|=|I_{\text{n}}|=G$

\STATE Compute rewards $r_i \leftarrow R(q,o_i)$ for all $i$;
$r_{\text{avg}} \leftarrow \frac{1}{2G}\sum_i r_i$

\STATE \textbf{(Rejection)} 
\IF{$r_{\text{avg}} > \tau$} 
    \STATE \ easy query: sampling $k=G/2$ samples
    \STATE $ \textbf{idx}_t \leftarrow \textbf{ArgSort}({O_t})$
    \STATE $ \mathcal{S}_t \leftarrow \{\textbf{idx}_t[j] : j=1,\ldots,k\}$
    \STATE $ \textbf{idx}_n \leftarrow \textbf{ArgSort}({O_n})$
    \STATE $ \mathcal{S}_n \leftarrow \{\textbf{idx}_n[j] : j=1,\ldots,k\}$
\ELSE
    \STATE \ hard query: keep all samples
\ENDIF

\STATE $\mathcal{S} \leftarrow \mathcal{S}_{\text{t}} \cup \mathcal{S}_{\text{n}}$ 
\STATE $m_i \leftarrow \mathbb{I}[\, i \in \mathcal{S}\,]$ for $i=1,\ldots,2G$

\STATE \textbf{return} $\mathcal{S},\, m$
\end{algorithmic}
\end{algorithm}

\begin{table*}[ht]
\small
\centering
\caption{Multimodal Dataset Information for Model Training and Evaluation}
\label{tab:data_composition}
\begin{tabular}{lccc}
\toprule
\multirow{2}{*}{Modality} & \multicolumn{3}{c}{Source Datasets} \\
\cmidrule(lr){2-4}
 & SFT & RL & Evaluation \\
\midrule
Text-only & 
\makecell[l]{Alpaca-gpt4 \cite{peng2023instruction} \\ OpenMath \cite{moshkov2025aimo2} \\  OpenThoughts\cite{guha2025openthoughtsdatarecipesreasoning}} & 
TextRL \cite{chen2025advancing} & 
\makecell[l]{mathQA \cite{mathqa} \\ ARC \citep{arc}} \\
\addlinespace[0.3em]
Text-Audio & 
Audio-Reasoner \cite{zhifei2025audio} & 
Audio-Reasoner & 
\makecell[l]{MMAR \cite{mmar} \\ MMAU \cite{mmau}} \\
\addlinespace[0.3em]
Text-Vision & 
\makecell[l]{Multimath \cite{peng2024multimath} \\ LLaVA-R1\cite{di_zhang_2025}} & 
Vision-R1 \cite{huang2025vision} & 
\makecell[l]{MathVision \cite{mathvision} \\ MMMU \cite{mmmu}} \\
\addlinespace[0.3em]
Text-Vision-Audio & 
Omni-instruct \cite{omnibench} & 
Omni-instruct \cite{omnibench} & 
OmniBench \cite{omnibench} \\
\midrule
Count & 571k & 10k & 3.6k \\
\bottomrule
\end{tabular}
\end{table*}

\section{Dataset}
In this section, we describe the construction of the datasets used for supervised fine-tuning (SFT), reinforcement learning (RL), and evaluation benchmarking.
\subsection{SFT Dataset}
As discussed in Section~\ref{sub_sec:adaptive sft}, our SFT dataset consists of two components:
(1) \textit{coarse-level adaptive data} and
(2) \textit{precise-level adaptive data}.
The coarse-level data is derived from existing reasoning and non-reasoning datasets, while the precise-level data is constructed using our designed strategy to classify samples into easy or hard problems.

\begin{table*}[ht]
\centering
\caption{Text-Audio Benchmark Accuracy and Thinking Rate Comparison. L1-L5 denotes the difficulty level. ACC denotes the Pass@1 Accuracy. Rate denotes the probability of using thinking mode. \textbf{Bold} for the best result and \underline{underline} for the second-best result.}
\label{tab:audio_benchmark}
\begin{tabular}{lcccccccccccc}
\toprule
\multirow{2}{*}{Model} & \multicolumn{2}{c}{L1} & \multicolumn{2}{c}{L2} & \multicolumn{2}{c}{L3} & \multicolumn{2}{c}{L4} & \multicolumn{2}{c}{L5} & \multicolumn{2}{c}{ALL} \\
\cmidrule(lr){2-3} \cmidrule(lr){4-5} \cmidrule(lr){6-7} \cmidrule(lr){8-9} \cmidrule(lr){10-11} \cmidrule(lr){12-13}
 & Acc & Rate & Acc & Rate & Acc & Rate & Acc & Rate & Acc & Rate & Acc & Rate \\
\midrule
\multicolumn{13}{l}{\textbf{Expert models}} \\
Qwen-Audio & 0.81 & 0.00 & 0.62 & 0.00 & 0.36 & 0.00 & 0.30 & 0.00 & 0.26 & 0.00 & 0.52 & 0.00 \\
\midrule
\multicolumn{13}{l}{\textbf{Omni models}} \\
Qwen2.5-Omni & \textbf{0.99} & 0.00 & 0.91 & 0.00 & 0.65 & 0.00 & 0.22 & 0.00 & 0.06 & 0.00 & 0.65 & 0.00 \\
Qwen3-Omni & 0.96 & 0.00 & 0.85 & 0.00 & 0.60 & 0.00 & \textbf{0.54} & 0.00 & 0.36 & 0.00 & \underline{0.72} & 0.00 \\
Ours & \underline{0.98} & 0.17 & \textbf{0.91} & 0.37 & \textbf{0.65} & 0.60 & \underline{0.49} & 0.69 & \textbf{0.38} & 0.71 & \textbf{0.73} & 0.47 \\
\bottomrule
\end{tabular}
\end{table*}

\begin{table*}[ht]
\centering
\caption{The results of Text-Audio-Vision Benchmark. L1-L5 denotes the difficulty level. ACC denotes the Pass@1 Accuracy. Rate denotes the probability of using thinking mode.}
\label{tab:text_audio_image_benchmark}
\begin{tabular}{lcccccccccccc}
\toprule
\multirow{2}{*}{Model} & \multicolumn{2}{c}{L1} & \multicolumn{2}{c}{L2} & \multicolumn{2}{c}{L3} & \multicolumn{2}{c}{L4} & \multicolumn{2}{c}{L5} & \multicolumn{2}{c}{ALL} \\
\cmidrule(lr){2-3} \cmidrule(lr){4-5} \cmidrule(lr){6-7} \cmidrule(lr){8-9} \cmidrule(lr){10-11} \cmidrule(lr){12-13}
 & Acc & Rate & Acc & Rate & Acc & Rate & Acc & Rate & Acc & Rate & Acc & Rate \\
\midrule
Qwen2.5-Omni & 0.86 & 0.00 & 0.61 & 0.00 & 0.42 & 0.00 & 0.32 & 0.00 & 0.21 & 0.00 & 0.48 & 0.00 \\
Qwen3-Omni & 0.77 & 0.00 & 0.66 & 0.00 & 0.56 & 0.00 & 0.51 & 0.00 & 0.36 & 0.00 & 0.57 & 0.00 \\
Ours & \textbf{0.92} & 0.16 & \textbf{0.86} & 0.24 & \textbf{0.65} & 0.35 & \textbf{0.61} & 0.30 & \textbf{0.46} & 0.23 & \textbf{0.69} & 0.25 \\
\bottomrule
\end{tabular}
\end{table*}

\noindent \textbf{Coarse-level adaptive data.}
Table~\ref{tab:data_composition} summarizes the composition of our coarse-level SFT data.
The dataset covers multiple modalities, including text-only, text–audio, text–vision, and text–vision–audio pairs.
Based on our preliminary experiments, we find that high-quality reasoning data plays a critical role in enhancing the model’s reasoning ability.
Therefore, we maintain a reasoning-to-non-reasoning ratio of 2:1 to ensure sufficient exposure to reasoning trajectories.

\noindent \textbf{Precise-level adaptive data.}
To construct the precise-level adaptive data, we follow the difficulty calibration protocol described in Section~\ref{sub_sec:benchmark} to assign accurate difficulty labels.
Specifically, we define difficulty levels L1–L2 as \textit{easy} problems and L3–L5 as \textit{hard} problems.
Note that if certain samples are categorized as hard problems but do not contain reasoning traces, we employ our internal models to generate the corresponding reasoning processes for these samples.
In this dataset, the proportion of thinking and no-thinking samples is balanced at 1:1.
\subsection{RL Dataset} \label{sub_sec:rl_data}
Similarly, we construct an RL dataset encompassing text-only, text–audio, text–vision, and text–vision–audio scenarios.
Unlike the SFT dataset, the RL dataset does not rely on explicit easy/hard labels or reasoning annotations.
Inspired by \cite{Polaris2025}, we introduce a \textit{data-filtering strategy} to enhance data quality of RL stage. 
Specifically, we employ the SFT model to sample multiple candidate responses for each query (eight outputs per query).
We then filter out samples with 100\% or 0\% pass rates, thereby discarding extremely easy or overly difficult problems.
This filtering ensures that the RL dataset focuses on moderately challenging cases that are most beneficial for adaptive reasoning training.

\subsection{Adaptive Reasoning Benchmark} \label{sub_sec:benchmark}
To comprehensively evaluate the adaptive reasoning capabilities of multimodal models, we introduce the \textbf{Omni Adaptive Reasoning Benchmark}.
This benchmark is designed to systematically assess models across diverse modalities and varying levels of problem complexity.
It encompasses four distinct modalities:
(1) \textbf{Text-only},
(2) \textbf{Text–Audio},
(3) \textbf{Text–Visual}, and
(4) \textbf{Text–Visual–Audio}.
Each sample within a modality is annotated with one of five difficulty levels (L1–L5), where L1 denotes the simplest and L5 represents the most complex tasks.

\noindent \textbf{Data Curation and Composition.}
Our benchmark aggregates, filters, and harmonizes data from multiple high-quality, publicly available datasets to ensure broad coverage of reasoning scenarios and modalities.
This design allows comprehensive evaluation of a model’s capacity to adapt reasoning depth according to input complexity.
The detailed composition and data sources are summarized in Table~\ref{tab:data_composition}.

\begin{table*}[ht]
\centering
\caption{Performance on Text-only Adaptive Benchmark. L1-L5 denotes the difficulty level. ACC denotes the Pass@1 Accuracy. Rate denotes the probability of using thinking mode.}
\label{tab:text_bench}
\begin{tabular}{lcccccccccccc}
\toprule
\multirow{2}{*}{Model} & \multicolumn{2}{c}{L1} & \multicolumn{2}{c}{L2} & \multicolumn{2}{c}{L3} & \multicolumn{2}{c}{L4} & \multicolumn{2}{c}{L5} & \multicolumn{2}{c}{ALL} \\
\cmidrule(lr){2-3} \cmidrule(lr){4-5} \cmidrule(lr){6-7} \cmidrule(lr){8-9} \cmidrule(lr){10-11} \cmidrule(lr){12-13}
 & Acc & Rate & Acc & Rate & Acc & Rate & Acc & Rate & Acc & Rate & Acc & Rate \\
\midrule
\multicolumn{13}{l}{\textbf{Expert models}} \\
Deep-seek R1-7B & 0.92 & 0.95 & 0.84 & 0.92 & 0.84 & 0.94 & 0.81 & 0.86 & 0.27 & 0.83 & 0.74 & 0.89 \\
AdaptThink & 0.93 & 0.93 & 0.89 & 0.92 & 0.87 & 0.90 & 0.79 & 0.80 & 0.30 & 0.81 & 0.76 & 0.87 \\
\midrule
\multicolumn{13}{l}{\textbf{Omni models}} \\
Qwen2.5-Omni & 0.60 & 0.31 & 0.45 & 0.30 & 0.38 & 0.23 & 0.25 & 0.26 & 0.07 & 0.23 & 0.35 & 0.26 \\
Qwen3-Omni & \textbf{0.91} & 0.0 & 0.86 & 0.0 & \textbf{0.88} & 0.0 & \textbf{0.76} & 0.0 & \textbf{0.65} & 0.0 & \textbf{0.81} & 0.0 \\
Ours & \underline{0.83} & 0.39 & \textbf{0.87} & 0.45 & \underline{0.67} & 0.52 & \underline{0.49} & 0.50 & \underline{0.48} & 0.53 & \underline{0.66} & 0.48 \\
\bottomrule
\end{tabular}
\end{table*}

\begin{table*}[ht]
\centering
\caption{The results of Text-Vision Benchmark. L1-L5 denotes the difficulty level. ACC denotes the Pass@1 Accuracy. Rate denotes the probability of using thinking mode.}
\label{tab:text_image_benchmark}
\begin{tabular}{lcccccccccccc}
\toprule
\multirow{2}{*}{Model} & \multicolumn{2}{c}{L1} & \multicolumn{2}{c}{L2} & \multicolumn{2}{c}{L3} & \multicolumn{2}{c}{L4} & \multicolumn{2}{c}{L5} & \multicolumn{2}{c}{ALL} \\
\cmidrule(lr){2-3} \cmidrule(lr){4-5} \cmidrule(lr){6-7} \cmidrule(lr){8-9} \cmidrule(lr){10-11} \cmidrule(lr){12-13}
 & Acc & Rate & Acc & Rate & Acc & Rate & Acc & Rate & Acc & Rate & Acc & Rate \\
\midrule
\multicolumn{13}{l}{\textbf{Expert models}} \\
Qwen2.5-VL & 0.82 & 0.24 & 0.62 & 0.38 & 0.48 & 0.37 & 0.41 & 0.45 & 0.32 & 0.40 & 0.52 & 0.37 \\
R-4B  & 0.20 & 0.55 & 0.16 & 0.57 & 0.21 & 0.56 & 0.23 & 0.59 & 0.20 & 0.56 & 0.20 & 0.57 \\
\midrule
\multicolumn{13}{l}{\textbf{Omni models}} \\
Qwen2.5-Omni & 0.85 & 0.00 & \textbf{0.65} & 0.00 & 0.58 & 0.00 & 0.37 & 0.00 & 0.25 & 0.00 & 0.52 & 0.00 \\
Qwen3-Omni & 0.86 & 0.00 & 0.64 & 0.00 & 0.52 & 0.00 & \textbf{0.57} & 0.00 & \textbf{0.54} & 0.00 & \textbf{0.62} & 0.00 \\
Ours & \textbf{0.88} & {0.52} & 0.61 & {0.67} & \textbf{0.59} & {0.69} & \underline{0.42} & {0.72} & \underline{0.32} & {0.64} & \underline{0.56} & {0.64} \\
\bottomrule
\end{tabular}
\end{table*}

\noindent \textbf{Difficulty Calibration Protocol.}
\label{sec:calibration}
A core contribution of our benchmark is a rigorous, model-based calibration strategy for assigning difficulty levels.
To establish an objective measure of problem complexity, we propose a hierarchical, model-based approach.
This calibration framework employs three tiers of models, with the performance of each tier evaluated over eight sampling runs per test instance.

\begin{itemize}
\item \textbf{M1 (Base Model):} A relatively weaker foundation model (Qwen-Omni-7B) serving as a baseline reference.
\item \textbf{M2 (Specialist Teacher Model):} High-performing, modality-specific models that are not explicitly trained for reasoning (e.g., Qwen2.5-7B for text, Qwen2.5-VL-7B for vision, Gemini Flash 2.0 for audio).
\item \textbf{M3 (Specialist Reasoning Model):} State-of-the-art reasoning-specialist models for each modality (e.g., DeepSeek-R1-7B for text, Seed-VL 1.5 for vision, and Gemini 2.5 Pro for audio).
\end{itemize}

Based on the relative performance of these three tiers, we define five distinct difficulty levels as follows:

\begin{itemize}
\item \textbf{L1 – Trivial:} A problem is classified as L1 if it is consistently solved by the base model (M1), achieving correct responses in more than six of eight sampling runs.
\item \textbf{L2 – Simple:} Problems are labeled L2 when M1 succeeds only occasionally (more than three correct runs) while the specialist teacher model (M2) achieves near-perfect accuracy (more than six correct runs).
\item \textbf{L3 – Intermediate:} Assigned when M1 largely fails (fewer than three correct runs) but M2 demonstrates consistent mastery (more than seven correct runs).
\item \textbf{L4 – Challenging:} Problems fall into L4 when M2 struggles (fewer than three correct runs), yet the specialist reasoning model (M3) consistently produces correct answers (more than six runs).
\item \textbf{L5 – Expert:} The highest difficulty level, consisting of problems that challenge even the advanced reasoning model (M3), which fails to produce correct responses in most runs (fewer than three correct).
\end{itemize}

This systematic calibration process establishes a well-defined complexity gradient across the benchmark.
As demonstrated in Section~\ref{sec:experiments}, our experiments validate the reliability of this classification, showing a clear monotonic relationship between increasing difficulty levels and decreasing model accuracy across a diverse set of open-source models.
\begin{table*}[t]
\centering
\caption{The ablation study of influence of SFT and RL.}
\label{tab:ablation_sft_rl}
\begin{tabular}{lcccccccc}
\toprule
\multirow{2}{*}{Model} & \multicolumn{2}{c}{Text-only} & \multicolumn{2}{c}{Text-Audio} & \multicolumn{2}{c}{Text-Visual} & \multicolumn{2}{c}{Text-Visual-Audio} \\
\cmidrule(lr){2-3} \cmidrule(lr){4-5} \cmidrule(lr){6-7} \cmidrule(lr){8-9}
 & Acc & Rate & Acc & Rate & Acc &  Rate & Acc & Rate \\
\midrule
Base & 0.35 & 0.26 & 0.64 & 0.00 & 0.52 & 0.00 & 0.48 & 0.00 \\
Base+SFT & 0.49 & 0.16 & 0.64 & 0.00 & 0.53 & 0.16 & 0.50 & 0.00 \\
Base + RL & 0.60 & 0.85 & 0.66 & 0.00 & 0.57 & 0.40 & 0.50 & 0.00 \\
Base+SFT+RL & 0.66 & 0.48 & 0.73 & 0.47 & 0.56 & 0.64 & 0.69 & 0.25 \\
\bottomrule
\end{tabular}
\end{table*}

\begin{table*}[ht]
\centering
\caption{The ablation study of Adaptive GRPO.}
\label{tab:ablation_grpo}
\begin{tabular}{l*{8}{c}}
\toprule
\multirow{2}{*}{Model} & \multicolumn{2}{c}{Text-only} & \multicolumn{2}{c}{Text-Audio} & \multicolumn{2}{c}{Text-Visual} & \multicolumn{2}{c}{Text-Visual-Audio} \\
\cmidrule(lr){2-3} \cmidrule(lr){4-5} \cmidrule(lr){6-7} \cmidrule(lr){8-9}
 & Acc & Rate & Acc & Rate & Acc & Rate & Acc & Rate \\
\midrule
GRPO-only & 0.82 & 1.0 & 0.64 & 0.0 & 0.55 & 0.0 & 0.53 & 0.0 \\
Adaptive GRPO (proposed) & 0.66 & 0.48 & 0.73 & 0.47 & 0.56 & 0.64 & 0.69 & 0.25 \\
\bottomrule
\end{tabular}
\end{table*}

\section{Experiments} \label{sec:experiments}

\subsection{Experimental Setting}

\noindent \textbf{Training Details.}
As described in Section~4, our proposed method consists of two stages: supervised fine-tuning (SFT) and reinforcement learning (RL).
For the SFT stage, we initialize the model with the pre-trained Qwen2.5-Omni-7B and train it on our collected SFT dataset for one epoch using a learning rate of $1\times10^{-5}$.
For the RL stage, we load the checkpoint from the SFT stage and continue training on the RL dataset for one epoch with a learning rate of $1\times10^{-6}$.

\noindent \textbf{Baselines.}
To evaluate the effectiveness of our proposed model, we compare it against several state-of-the-art (SOTA) baselines.
These include omni-modal models such as Qwen2.5-Omni-7B and Qwen3-Omni-30B, as well as domain-specific expert or adaptive reasoning models across different modalities.
For the text-only benchmark, we compare with AdaptThink~\citep{adaptthink} and DeepSeek-R1-7B~\citep{deepseek-r1}.
For the text–vision benchmark, we include Qwen2.5-VL-7B~\citep{qwen2.5-vl} and R-4B~\citep{r-4b} as representative baselines.
Additional comparisons are also made with Qwen2.5-Audio in the text–audio domain to ensure comprehensive evaluation across modalities.

\noindent \textbf{Evaluation Metrics.}
Since the benchmark tasks are formulated as multiple-choice problems, we adopt \textit{Pass@1 Accuracy} as the primary metric for assessing model performance.
To further evaluate adaptive reasoning behavior, we also report the \textit{thinking rate}, defined as the proportion of responses containing reasoning traces.
Ideally, an adaptive reasoning model should exhibit a higher thinking rate on difficult problems and a lower one on easier problems, demonstrating its ability to dynamically adjust reasoning depth according to task complexity.

\subsection{Main Results}

\noindent \textbf{Comparison with Previous SOTA Models.}
We first evaluate our method on the \textbf{Omni Adaptive Reasoning Benchmark}.
Experimental results are presented in Table~\ref{tab:audio_benchmark} through Table~\ref{tab:text_image_benchmark}.
We summarize the key observations as follows:

\noindent (1) \textbf{Accuracy Performance.}
Compared to the base model (Qwen2.5-Omni-7B), our proposed method achieves substantial improvements across all benchmark settings.
Notably, our model even surpasses the larger Qwen3-Omni-30B model on the text–audio, text–vision, and text–audio–vision benchmarks.
We observe that Qwen3-Omni exhibits slightly higher accuracy on the text-only benchmark, which we attribute to its larger parameter scale and stronger text understanding capability compared to our 7B model.

\noindent (2) \textbf{Adaptive Reasoning Behavior.}
From the perspective of adaptive reasoning, our model demonstrates a clear correlation between reasoning frequency and problem difficulty:
it exhibits a higher thinking rate on hard problems (L3–L5) and a lower rate on easy problems (L1–L2).
This trend confirms that our model effectively learns to adjust reasoning depth according to task complexity, validating the core motivation of the proposed adaptive reasoning framework.
\begin{table*}[ht]
\centering
\caption{Performance Comparison of Different SFT Data Scales.}
\label{tab:ablation_sft_data}
\begin{tabular}{l*{8}{c}}
\toprule
\multirow{2}{*}{Setting} & \multicolumn{2}{c}{Text-only} & \multicolumn{2}{c}{Text-Audio} & \multicolumn{2}{c}{Text-Visual} & \multicolumn{2}{c}{Text-Visual-Audio} \\
\cmidrule(lr){2-3} \cmidrule(lr){4-5} \cmidrule(lr){6-7} \cmidrule(lr){8-9}
 & Acc & Rate & Acc & Rate & Acc & Rate & Acc & Rate \\
\midrule
Small-scale SFT  & 0.39 & 0.12 & 0.59 & 0.00 & 0.26 & 0.00 & 0.50 & 0.00 \\
Large-scale SFT & 0.48 & 0.20 & 0.60 & 0.00 & 0.51 & 0.10 & 0.51 & 0.00 \\
All SFT & 0.49 & 0.16 & 0.64 & 0.00 & 0.53 & 0.16 & 0.50 & 0.00 \\
% \midrule

% All SFT + Adaptive GRPO & 0.66 & 0.48 & 0.73 & 0.47 & 0.56 & 0.64 & 0.69 & 0.25 \\
\bottomrule
\end{tabular}
\end{table*}

\begin{figure}[t]
    \centering
    \includegraphics[width=0.5\textwidth, height=0.35\textwidth]{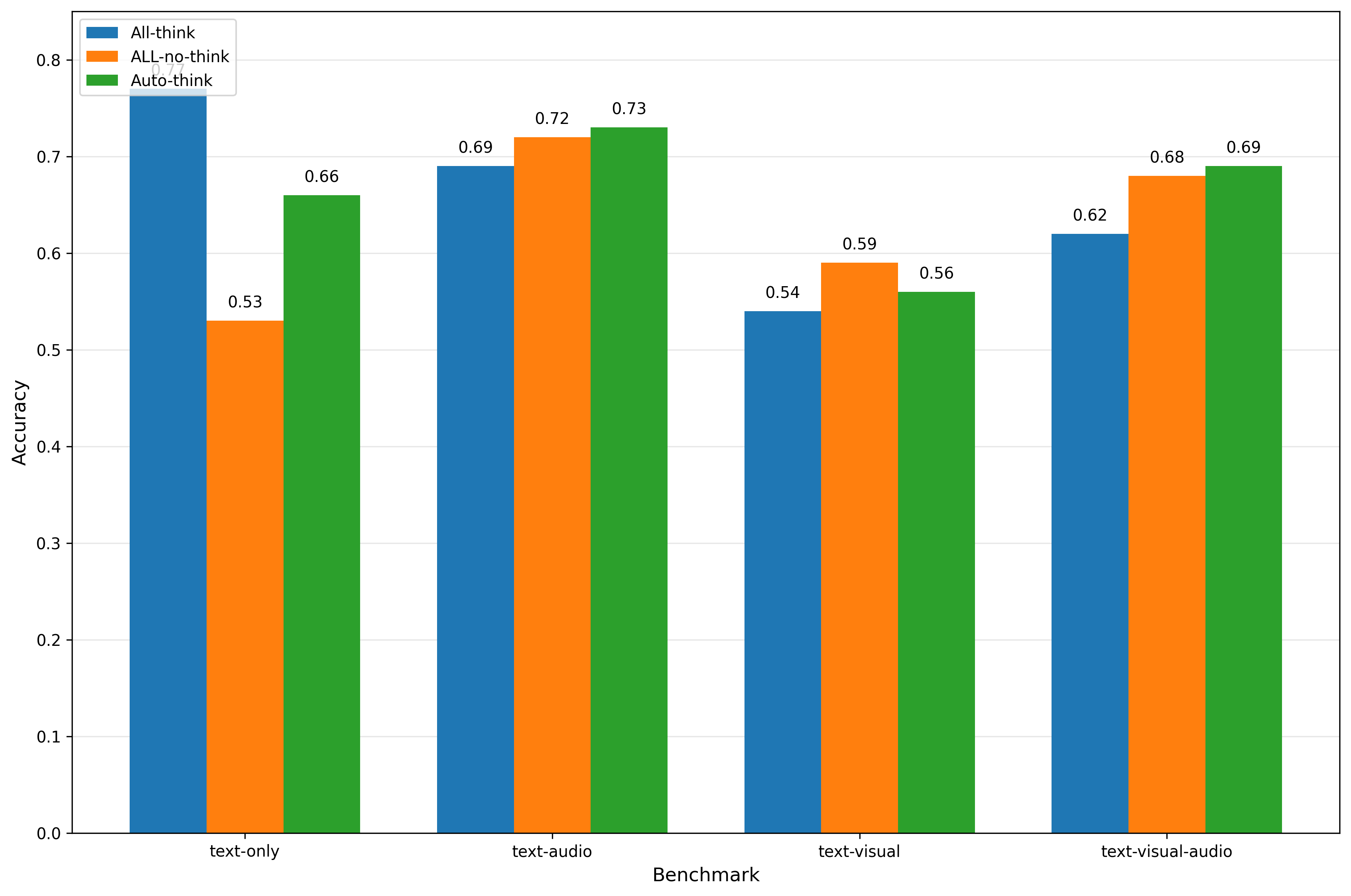}
    \caption{The performance comparison between adaptive think, all-think, and no-think.}
    \label{fig:three-mode}
\end{figure}
\begin{table*}[ht]
\centering
\caption{Performance Comparison of Different RL data.}
\label{tab:ablation_rl_data}
\begin{tabular}{l*{8}{c}}
\toprule
\multirow{2}{*}{Setting} & \multicolumn{2}{c}{Text-only} & \multicolumn{2}{c}{Text-Audio} & \multicolumn{2}{c}{Text-Visual} & \multicolumn{2}{c}{Text-Visual-Audio} \\
\cmidrule(lr){2-3} \cmidrule(lr){4-5} \cmidrule(lr){6-7} \cmidrule(lr){8-9}
 & Acc & Rate & Acc & Rate & Acc & Rate & Acc & Rate \\
\midrule
No-data-filter & 0.60 & 0.55 & 0.71 & 0.37 & 0.52 & 0.32 & 0.64 & 0.22 \\
RL-filter strategy & 0.66 & 0.48 & 0.73 & 0.47 & 0.56 & 0.64 & 0.69 & 0.25 \\
\bottomrule
\end{tabular}
\end{table*}

\noindent \textbf{Comparison with All-Thinking and No-Thinking.}
In this experiment, we evaluate the performance of our adaptive reasoning framework against two extreme settings: \textit{all-thinking} and \textit{no-thinking} modes.
Following the configuration of Qwen3~\citep{qwen3}, we insert special control tokens into the prompt to explicitly direct the model to adopt either reasoning or non-reasoning behavior.
Figure~\ref{fig:three-mode} presents the performance comparison among the three reasoning styles.
We observe the following:
(1) the adaptive reasoning mode consistently outperforms the no-thinking mode across all benchmarks, demonstrating the importance of adaptive reasoning; and
(2) compared with the all-thinking mode, the adaptive reasoning mode achieves better overall performance on most benchmarks.
Interestingly, the all-thinking mode performs slightly better on the text-only benchmark.
We hypothesize that this may result from the large-scale, high-quality text reasoning data introduced during the SFT stage, which enhances text-domain reasoning ability.

\subsection{Ablation Study}

To obtain a more comprehensive understanding of the proposed adaptive reasoning framework, we conduct systematic ablation studies, isolating the contribution of each component under controlled experimental settings with identical architectures and hyperparameters.

\noindent \textbf{Influence of SFT and RL.}
Our adaptive reasoning framework consists of two core stages: SFT and RL.
To evaluate their effectiveness, we conduct ablation experiments for each stage, as shown in Table~\ref{tab:ablation_sft_rl}.
We make the following observations:
(1) both SFT and RL contribute positively to overall accuracy;
(2) applying SFT alone cannot yield adaptive reasoning behavior, highlighting the essential role of the proposed Adaptive GRPO stage; and
(3) removing the SFT stage results in suboptimal performance during RL, confirming that SFT provides a crucial foundation for subsequent policy optimization.

\noindent \textbf{Ablation of Adaptive GRPO.}
We further examine the effectiveness of our proposed Adaptive GRPO algorithm.
As shown in Table~\ref{tab:ablation_grpo}, compared to the vanilla GRPO baseline, the model trained with Adaptive GRPO successfully learns adaptive reasoning behavior, achieving significantly better performance across difficulty levels.

\noindent \textbf{Influence of SFT Data.}
We next analyze the impact of SFT data composition.
As described in Section~4, our SFT dataset comprises two components:
(1) a small-scale, precise-level dataset designed to help the model differentiate easy and hard problems, and
(2) a large-scale, coarse-level dataset providing broad coverage of reasoning and non-reasoning examples.
Table~\ref{tab:ablation_sft_data} shows that using only one component leads to suboptimal performance, whereas combining both datasets yields the best results.
This finding demonstrates the effectiveness of our two-tier SFT data construction strategy.

\noindent \textbf{Influence of the RL Data Filtering Strategy.}
As described in Section \ref{sub_sec:rl_data}, we introduce a data-filtering strategy to improve data quality during the RL stage.
To validate its effectiveness, we conduct ablation experiments comparing models trained with and without this filtering mechanism.
As shown in Table~\ref{tab:ablation_rl_data}, incorporating the proposed filtering strategy consistently improves final model performance, indicating that removing extremely easy or overly difficult samples helps stabilize reinforcement learning and focus optimization on informative cases.

\section{Conclusion}
In this paper, we proposed Omni-AutoThink, a unified framework that enables multimodal models to perform adaptive reasoning—deciding when and how deeply to think based on task difficulty.
Our method combines Adaptive SFT and Adaptive GRPO, jointly enhancing reasoning capability and decision control.
We also introduced the Omni Adaptive Reasoning Benchmark, covering four modalities and five calibrated difficulty levels.
Extensive experiments show that Omni-AutoThink significantly improves both reasoning accuracy and adaptivity over previous state-of-the-art models.

\bibliography{custom}

\appendix

\section{Appendix}
\label{sec:appendix}

\subsection{Adaptive Reasoning Prompt} \label{appendix:prompt_design}
In the following, we show the prompts designed for Omni models. The first prompt asks the model to directly output the answer, while the second encourages the model to perform adaptive reasoning.
\begin{center}
\begin{minipage}{0.5\textwidth}
\fboxrule=2pt
\fboxsep=10pt
\colorbox{lightblue}{\parbox{\dimexpr\textwidth-2\fboxsep-2\fboxrule}{
    \textcolor{bluebar}{\textbf{\large Base Prompt}}\\[5pt]
    You are a virtual human system, capable of perceiving auditory and visual inputs, as well as generating text and speech.
}}
\end{minipage}
\end{center}

\begin{center}
\begin{minipage}{0.5\textwidth}
\fboxrule=2pt
\fboxsep=10pt
\colorbox{lightblue}{\parbox{\dimexpr\textwidth-2\fboxsep-2\fboxrule}{
    \textcolor{bluebar}{\textbf{\large Adaptive Prompt}}\\[5pt]
You are a virtual human system, capable of perceiving auditory and visual inputs, as well as generating text and speech. Your primary goal is to analysis and solve user's question.
    First, identify whether this problem requires thinking. If the problem requires thinking, output the thinking process in <think>\texttt{\textbackslash n} reasoning step \texttt{\textbackslash n}</think> and final answer inside <answer> </answer>. 
    If no thinking is required, please output <think>\textbackslash n\textbackslash n</think> to denote the empty thinking process, then output answer in <answer> </answer>. The Assistant is encouraged to use 
    the <answer></answer> tag whenever possible, while ensuring accuracy.
}}
\end{minipage}
\end{center}

\end{document}